\documentclass[runningheads]{llncs}
\usepackage{bm}
\usepackage{amsmath}
\usepackage{threeparttable}
\usepackage{multirow}
\usepackage{graphicx}
\usepackage{url}
\graphicspath{{img/}}
\DeclareGraphicsExtensions{.png,.pdf}
\usepackage[caption=false,font=small]{subfig}
\begin{document}
\title{A Fine-Grain Error Map Prediction and Segmentation Quality Assessment Framework for Whole-Heart Segmentation}
\titlerunning{Error Map Prediction and Quality Assessment for WHS}
%
\author{Rongzhao Zhang \and Albert C.S. Chung 
}
%
%
\institute{The Hong Kong University of Science and Technology, Hong Kong, China
\email{\{rzhangbe,achung\}@cse.ust.hk}
}

\maketitle              
\begin{abstract}
When introducing advanced image computing algorithms, e.g., whole-heart segmentation, into clinical practice, a common suspicion is how reliable the automatically computed results are. In fact, it is important to find out the failure cases and identify the misclassified pixels so that they can be excluded or corrected for the subsequent analysis or diagnosis. However, it is not a trivial problem to predict the errors in a segmentation mask when ground truth (usually annotated by experts) is absent. In this work, we attempt to address the pixel-wise error map prediction problem and the per-case mask quality assessment problem using a unified deep learning (DL) framework. Specifically, we first formalize an error map prediction problem, then we convert it to a segmentation problem and build a DL network to tackle it. We also derive a quality indicator (QI) from a predicted error map to measure the overall quality of a segmentation mask. To evaluate the proposed framework, we perform extensive experiments on a public whole-heart segmentation dataset, i.e., \textit{MICCAI 2017 MMWHS}. By 5-fold cross validation, we obtain an overall Dice score of 0.626 for the error map prediction task, and observe a high Pearson correlation coefficient (PCC) of 0.972 between QI and the actual segmentation accuracy (Acc), as well as a low mean absolute error (MAE) of 0.0048 between them, which evidences the efficacy of our method in both error map prediction and quality assessment.

\end{abstract}
\section{Introduction}
Assessing per-case image segmentation quality is an important issue when researchers want to develop a computer-aided diagnosis (CADx) system or integrate automated image analysis methods into large-scale medical studies. Since image segmentation usually serves as a low-level module in a CADx system or a clinical study pipeline, errors incurred by segmentation algorithms will be delivered or even amplified in the subsequent calculation of image-based measurements and other downstream procedures, which may result in misleading statistical conclusions or slow down the diagnosis process. Automatic quality assessment is an appealing solution to such problems, which, ideally, should not only report the per-case segmentation quality, but also highlight those misclassified pixels, so that doctors (or an automatic system) can easily verify the reliability of a segmentation result and decide whether to keep it for further analysis. Besides, an automatic pixel-wise error prediction algorithm has a great potential in medical training, where it can provide inexperienced students with fine-grain feedback by pointing out which pixels are mislabeled.

Although the quality assessment for segmentations can be done by simply comparing with experts' annotation, this method is too costly to be applied in large-scale studies or automated pipelines. In natural image analysis area, there have been a number of unsupervised image segmentation evaluation methods \cite{zhang2008image}, which employ low-level features, e.g., color error, texture, entropy, and their combinations to measure a segmentation's visual consistency with human observers. However, the application of such methods in medical area remains unclear \cite{valindria2017reverse}. Reserve validation method \cite{zhong2010cross} trains classifiers with pseudo ground truth to quantify how well a classifier performs on a target domain, but it can only give a single quality measurement for a classifier across the whole test set, which cannot meet the per-case demand of a quality assessment algorithm. Recently, Valindria \textit{et al.} proposed reverse classification accuracy (RCA) \cite{valindria2017reverse,robinson2017automatic} method which is able to evaluate the quality for each single case, but this method has high computational cost and cannot predict a fine-grain error map. Robinson \textit{et al.} developed a deep learning model to directly regress the Dice Similarity Coefficient (DSC) of a segmentation mask, which is much faster but still can only provide an image-level measurement and requires a large-scale training set.

In this study, we build an automatic quality assessment framework that is capable of simultaneous pixel-wise and per-case evaluation for segmentation masks. Specifically, we first formally define the pixel-wise error map prediction problem, and then show the capacity of a modern deep learning (DL) model in predicting error maps for auto-generated segmentation masks. We also derive a quality indicator (QI) from the output error maps, which can measure segmentation quality in a per-case manner. To generate diverse and representative segmentation masks, we train a VoxResNet \cite{chen2018voxresnet} and its 2D version on the training sets, and collect all their side (induced by the deep supervision \cite{lee2015deeply} paths) and final outputs as sample segmentations. To demonstrate the efficacy of our method, we evaluate it on a public 3D whole-heart segmentation dataset, i.e., \textit{MICCAI 2017 MMWHS}. The quality and quantity results of a 5-fold cross validation show that our framework is able to identify the misclassified pixels in an input mask with satisfiable accuracy. We also observe a strong correlation between QI and the actual segmentation accuracy (Acc), as well as between QI and DSC score, evidencing the capacity of our framework working as an image-level segmentation quality evaluator. To the best of our knowledge, this is the first time that the segmentation quality assessment problem is addressed in a pixel-wise manner for medical images, and we are also a pioneer who manages to predict per-case segmentation quality accurately only based on relatively small training sets (e.g., 16 training MRI scans in each fold).

\section{Method}
\begin{figure}[]
	\subfloat[Error Map Prediction Framework]{\includegraphics[width=0.54\textwidth]{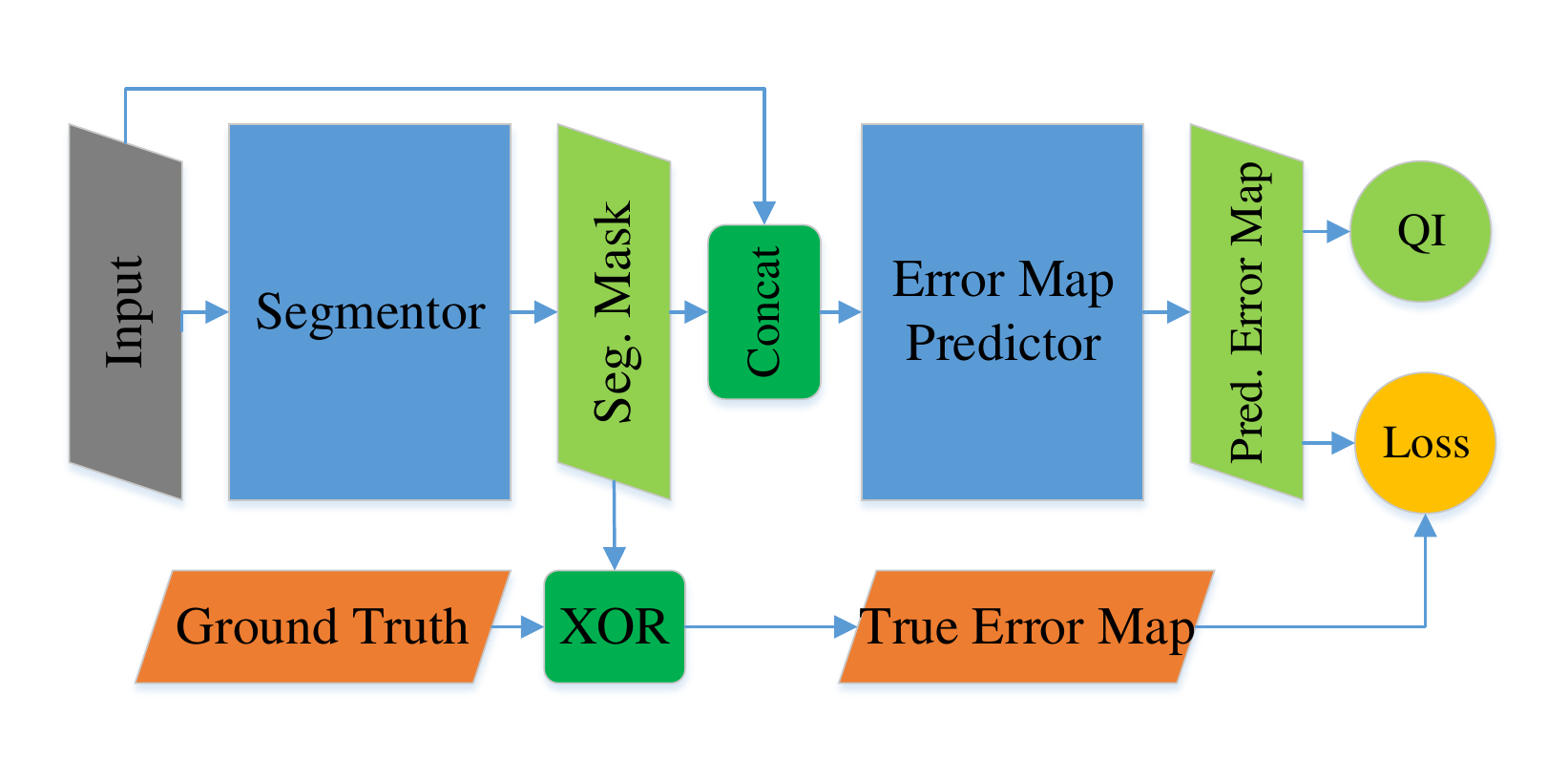}} %
	\hspace{-0.1in}
	\subfloat[DL Model Architecture]{\includegraphics[width=0.46\textwidth]{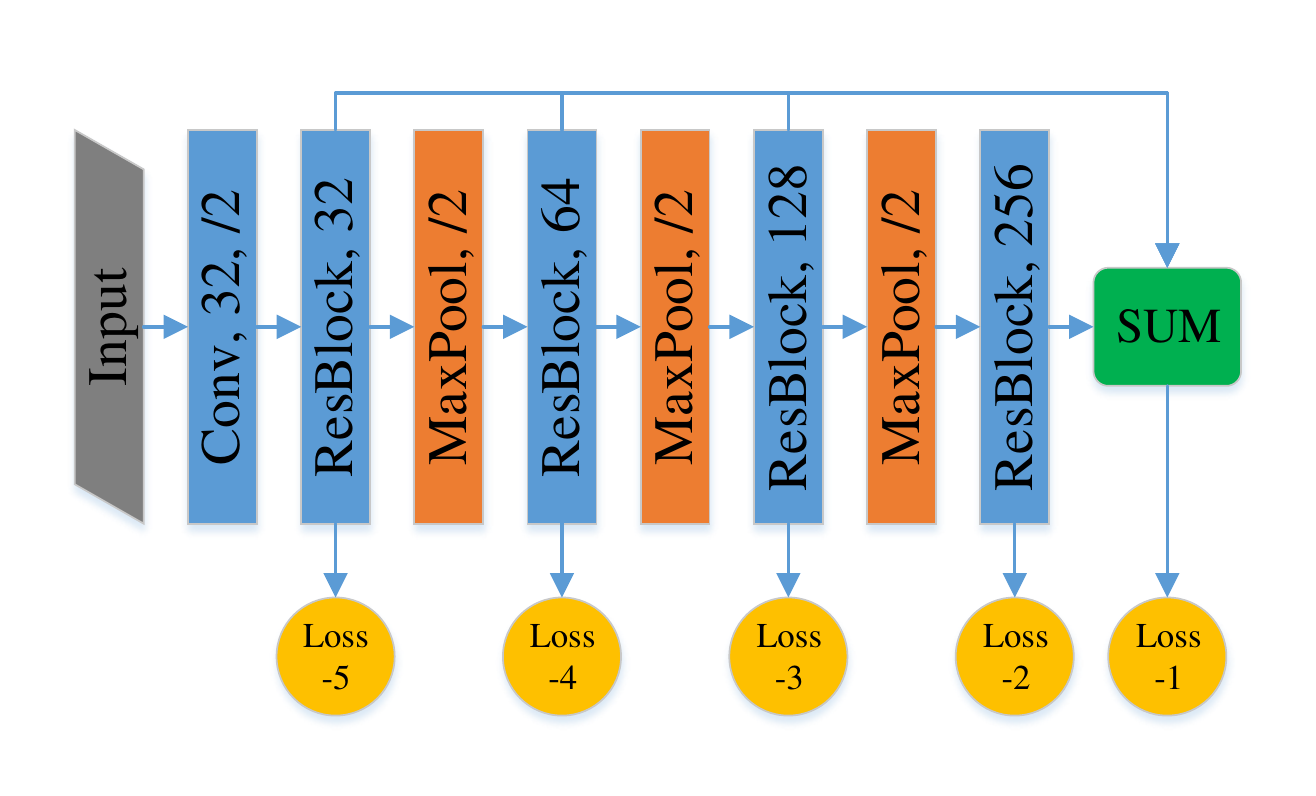}} %
\caption{Flowchart of the proposed error map prediction framework and the DL model architecture (VoxResNet) employed in this paper.} \label{fig_archi}
\end{figure}

Our method is mainly composed of a mask generation part (segmentor) and an error map prediction part (error map predictor), as shown in Fig. \ref{fig_archi}(a). In this section, we will first define the error map prediction problem, then elaborate the mask generation and error map prediction methods, and finally detail the training of the proposed framework.

\subsection{Formulation of Error Map Prediction Problem}
\begin{figure}[t]
	\subfloat[Segmentation Error Map]{\includegraphics[width=0.42\textwidth]{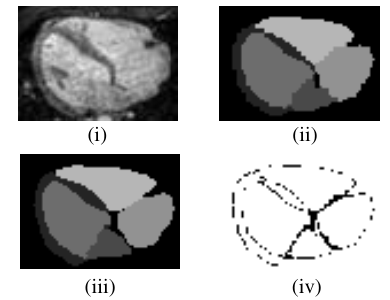}} %
	\hspace{0.1in}
	\subfloat[DSC Histogram]{\includegraphics[width=0.57\textwidth]{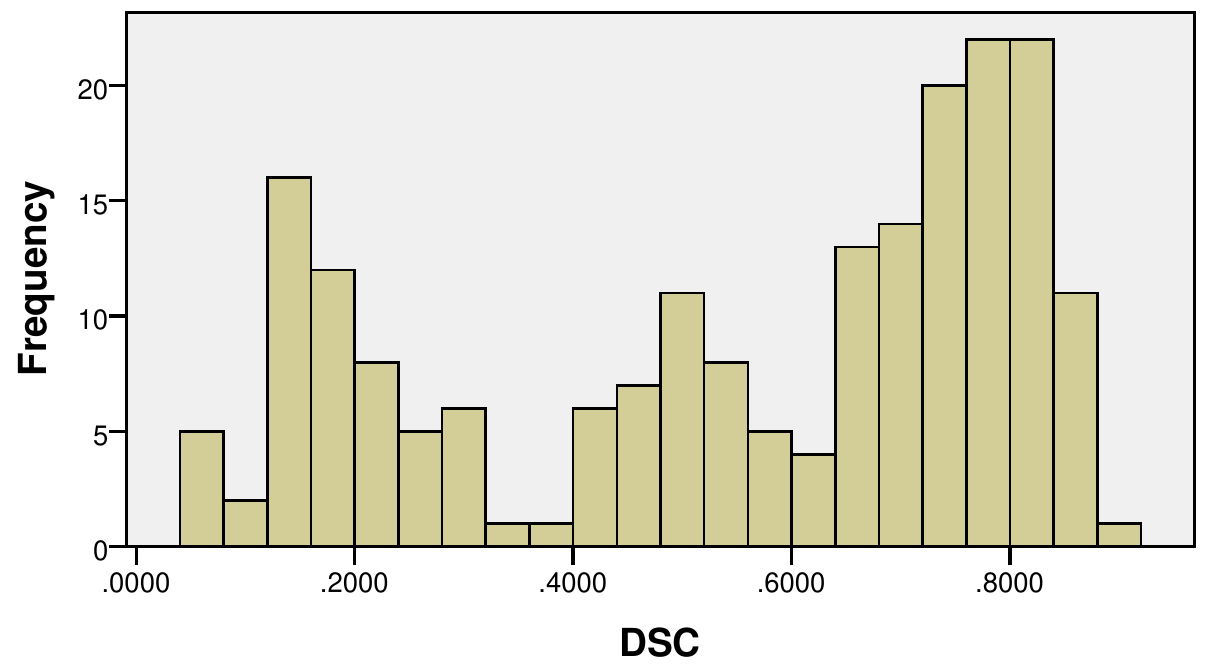}} %
	\caption{(a) An example error map, where (i), (ii), (iii) and (iv) are the original image, auto-generated mask, ground truth and error map, respectively. Note that the intensity of the error map is reversed (0 for error and 1 for correct pixels) for better visualization. (b) Histogram of DSC score of the generated segmentation masks.} \label{fig_hist_dsc}
\end{figure}

We define the error map $\mathcal{E}$ of a segmentation mask $S$ as
\begin{equation}\label{eq_em1}
  \mathcal{E}(i) = \begin{cases}1, &S(i)\neq GT(i), \cr 0, &S(i)=GT(i),\end{cases}
\end{equation}
where $GT$ is the ground truth segmentation and $i$ specifies the pixel (voxel) location. $S(i), GT(i)\in \{0,1,\cdots,C\}$, where $C$ is the number of foreground classes and 0 denotes the background class. When ground truth segmentation is not available, we build a model $M$ that is parameterized by $\theta$ to estimate the error map:
\begin{equation}\label{eq_em2}
\widehat{\mathcal{E}} = M(I,S;\theta),
\end{equation}
where $I$ denotes the original image, $\widehat{\mathcal{E}}$ is the predicted error map for the segmentation mask $S$. Thus, given a dataset $\mathcal{D}=\{I_i,\{S_i^k\}_{k=1}^{m},GT_i\}_{i=1}^N$, the error map prediction problem can be formulated as an optimization task over model parameter $\theta$:
\begin{equation}
\min_\theta \frac{1}{mN}\sum_{i=1}^{N}\sum_{k=1}^{m}d\left(\mathcal{E}(S_i^k,GT_i), \widehat{\mathcal{E}}(I_i,S_i^k;\theta)\right),
\end{equation}
where $N$ is the number of images, $m$ is the number of generated segmentation masks for each image, $d(\cdot,\cdot)$ is a distance metric such as cross entropy, which measures the difference between the true and the predicted error maps. An example error map is shown in Fig. \ref{fig_hist_dsc}(a).

\subsection{Mask Generation}
To enable the training and evaluation of the error map predictor, we train two different CNN models on the training sets and collect all their outputs to form a mask set $\{S_i^k\}_{i,k}$. One segmentation model is VoxResNet \cite{chen2018voxresnet}, a representative and state-of-the-art CNN model designed for volumetric medical image segmentation tasks, which leverages residual connections \cite{he2016deep} and combines multi-scale features to make a quality prediction, as schematically illustrated in Fig. \ref{fig_archi}(b). To encourage the diversity of collected masks, we also employ a 2D version of VoxResNet to perform the generation work, as it is believed that the outputs of 2D and 3D models can be significantly different since their receptive fields are distinctive. Further, the side outputs of these models, which are generated by the deep supervision pathways (corresponding to Loss-2 to Loss-5 in Fig. \ref{fig_archi}(b)), are also added to the mask collection to involve more examples with various segmentation qualities. Overall, the number of auto-generated masks for each scan is $m= 2\times (4+1)=10$, where 4 is the number of side outputs in a segmentor. The DSC score histogram of the generated segmentations is shown in Fig. \ref{fig_hist_dsc}(b).

\subsection{Error Map Predictor and Quality Indicator}
Since an error map has the same size as the corresponding segmentation mask and takes 0-1 values, the error map predictor can be implemented by a binary segmentation model. Without loss of generality, we employ another VoxResNet to carry out the prediction, which has the same architecture as the 3D segmentor but different input and output channels. As shown in Fig. \ref{fig_archi}(a), the error map predictor takes the concatenation of the segmentation mask (with one-hot coding) and the original image as inputs, and output a probabilistic map indicating the probability of each pixel being misclassified by the segmentor (then we can get the binary error map by thresholding). Since the mean value across a true error map is exactly the segmentation accuracy, we derive a quality indicator (QI) by averaging the predicted binary error map to measure the overall quality of the input mask (denoted by the QI node following the predicted error map in Fig. \ref{fig_archi}(a)). The training procedure for the predictor is similar to a standard segmentation model, with the difference that we generate its input masks on the fly by the segmentors (with fixed parameters) to save RAM space.

\subsection{Training Details} 
For the training of the segmentors, we optimize the summation of a cross entropy (CE) loss and a multi-class Dice loss on the training set with a standard segmentation pipeline. Then we apply the trained segmentors to both training and test sets, and collect their outputs on the training set for the subsequent training of the predictor, and save the outputs on the test set for evaluation of the error map predictor. In this way, both the segmentors and predictor can only access the training set during the learning phase, and their performance are evaluated solely on the test set. Besides, since the predictor is essentially a binary segmentation model, we optimize it by the summation of a CE loss and a binary Dice loss (error pixels as the positive class).

\section{Experiments and Results}
\subsubsection{Dataset}
We evaluated our framework on a public whole-heart segmentation dataset, i.e., \textit{MICCAI 2017 MMWHS}\footnote{\url{http://www.sdspeople.fudan.edu.cn/zhuangxiahai/0/mmwhs17/index.html}} \cite{zhuang2016multi}. We employed the 20 MRI scans in this dataset which are paired with manual annotations with 7 foreground classes. For preprocessing, we resampled each scans to an isotropic voxel resolution of $2\times 2\times 2$ mm, normalized their intensity to $[-1,1]$ and then extracted the heart region with a method similar to \cite{payer2017multi}. The preprocessed scans have a size of around $120\times 120\times 90$. Standard data augmentation methods are applied during the training of segmentors and the predictor, including random cropping (to a patch size of $96\times 96\times 80$, or $96\times 96$ for 2D), flipping along each axis and scaling. For all experiments, we run 5-fold cross validations since the dataset is relatively small.

\subsubsection{Metrics}
As the error map prediction problem is essentially a segmentation task, we employ common segmentation metrics to measure its performance, which include Dice similarity coefficient (DSC), accuracy (Acc), precision (Prec) and recall (Recl). Note that in the metric calculation we regard the error pixels as the positive class, though in the figures we show error pixels by low intensity for the sake of clear visualization.

\subsubsection{Error Map Predictor}
To thoroughly investigate the performance of the error map predictor, we evaluate it on different kinds of segmentation masks and report their performance in Table \ref{tab_em}. The predictor's performance on the final outputs of the 3D VoxResNet is tagged by `3D-Final', and its side outputs are tagged by `3D-2' to `3D-5', respectively. Ground truth masks are referred to as `GT'. We omit the detailed mask categories of 2D generators as they are similar to the 3D case, and only report their average values to save space. The last two rows show the predictor's overall performance on all masks with or without GT, respectively. We also list segmentation metrics in the table (last two columns) in order for better interpretation of the prediction results. Considering both prediction and segmentation metrics, we find that the error map predictor performs worse on those masks with better quality, and vice versa. This is because, for masks with good quality, the error regions are usually small (or thin) and near to the class boundaries (as the first case shown in Fig. \ref{fig_quali}), where it is hard to tell which pixel is wrong if GT is not available. On the other hand, low-quality masks tend to have larger error regions that do not concentrate on subtle boundaries, which can be easily identified by our error prediction model (an example is illustrated in the second row of Fig. \ref{fig_quali}). Another observation is that our error map predictor performs well for GT masks, on which the it achieves the highest prediction accuracy, i.e., 0.9967, meaning that our model can `feel' that a GT mask is of high quality, even though it has never seen this GT mask before. Note that the prediction DSC for GT masks is low because the true error map for a GT mask is all zero, such that even a single wrong pixel will lead to a zero DSC score. Overall, our error map predictor achieves a good DSC of 0.626 (0.592 if considering GT masks), demonstrating the efficacy of the proposed error map prediction model. Qualitative results can be found in Fig. \ref{fig_quali}, where several representative error map predictions are present.

\begin{table}[t]
\centering
\caption{Error Map Prediction Performance}
\label{tab_em}\setlength{\tabcolsep}{1.65mm}{
\begin{tabular}{lccccccc}\hline
	Mask Type    & \#Masks & DSC    & Acc    & Prec   & Recl   & Seg. DSC & Seg. Acc \\ \hline
	3D Final     & 20      & 0.4313 & 0.9757 & 0.3766 & 0.5479 & 0.8102   & 0.9816   \\
	3D-2         & 20      & \textbf{0.8611} & 0.9842 & 0.8132 & 0.9175 & 0.2751   & 0.938    \\
	3D-3         & 20      & 0.5666 & 0.9769 & 0.8133 & 0.6665 & 0.7781   & 0.9762   \\
	3D-4         & 20      & 0.5772 & 0.9676 & 0.4953 & 0.7074 & 0.6781   & 0.9674   \\
	3D-5         & 20      & 0.7421 & 0.9597 & 0.6501 & 0.8668 & 0.2387   & 0.932    \\
	GT           & 20      & 0.25   & \textbf{0.9967} & 0.25   & 1      & 1        & 1        \\ \hline
	3D-Average   & 100     & 0.6357 & 0.9728 & 0.6297 & 0.7412 & 0.5560   & 0.9590   \\
	2D-Average   & 100     & 0.6161 & 0.9691 & 0.5092 & 0.7927 & 0.5483   & 0.9647   \\ \hline
	Overall-auto & 200     & 0.6259 & 0.9710 & 0.5695 & 0.7670 & 0.5522   & 0.9619   \\
	Overall-GT   & 220     & 0.5917 & 0.9733 & 0.5404 & 0.7881 & 0.5929   & 0.9653   \\ \hline
\end{tabular}}
\end{table}

\newcommand{\picwidthVII}{0.76in}
\begin{figure*}[t]
	\captionsetup[subfigure]{labelformat=empty}
	\centering
	\subfloat{\includegraphics[width=\picwidthVII]{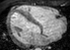}}
	\hspace{0in}	
	\subfloat{\includegraphics[width=\picwidthVII]{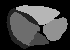}}
	\hspace{0in}	
	\subfloat{\includegraphics[width=\picwidthVII]{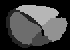}}
	\hspace{0in}	
	\subfloat{\includegraphics[width=\picwidthVII]{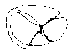}}
	\hspace{0in}	
	\subfloat{\includegraphics[width=\picwidthVII]{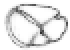}}
	\hspace{0in}	
	\subfloat{\includegraphics[width=\picwidthVII]{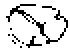}}
	
	\subfloat{\includegraphics[width=\picwidthVII]{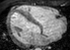}}
	\hspace{0in}	
	\subfloat{\includegraphics[width=\picwidthVII]{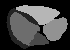}}
	\hspace{0in}	
	\subfloat{\includegraphics[width=\picwidthVII]{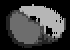}}
	\hspace{0in}	
	\subfloat{\includegraphics[width=\picwidthVII]{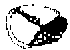}}
	\hspace{0in}	
	\subfloat{\includegraphics[width=\picwidthVII]{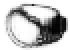}}
	\hspace{0in}	
	\subfloat{\includegraphics[width=\picwidthVII]{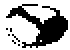}}
	
	\subfloat{\includegraphics[width=\picwidthVII]{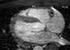}}
	\hspace{0in}	
	\subfloat{\includegraphics[width=\picwidthVII]{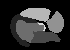}}
	\hspace{0in}	
	\subfloat{\includegraphics[width=\picwidthVII]{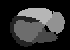}}
	\hspace{0in}	
	\subfloat{\includegraphics[width=\picwidthVII]{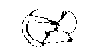}}
	\hspace{0in}	
	\subfloat{\includegraphics[width=\picwidthVII]{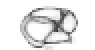}}
	\hspace{0in}	
	\subfloat{\includegraphics[width=\picwidthVII]{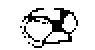}}
	
	\subfloat[MRI]{\includegraphics[width=\picwidthVII]{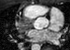}}
	\hspace{0in}	
	\subfloat[GT]{\includegraphics[width=\picwidthVII]{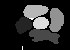}}
	\hspace{0in}	
	\subfloat[Seg]{\includegraphics[width=\picwidthVII]{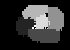}}
	\hspace{0in}	
	\subfloat[True EM]{\includegraphics[width=\picwidthVII]{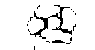}}
	\hspace{0in}	
	\subfloat[Soft EM]{\includegraphics[width=\picwidthVII]{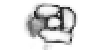}}
	\hspace{0in}	
	\subfloat[Pred EM]{\includegraphics[width=\picwidthVII]{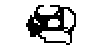}}
	
	\caption{Representative error map predictions. The last two columns show the raw (probabilistic) predicted error maps and the thresholded binary maps, respectively.}
	\label{fig_quali}
\end{figure*}

\subsubsection{Quality Indicator}
\begin{figure}[t]
	\subfloat[]{\includegraphics[width=0.5\textwidth]{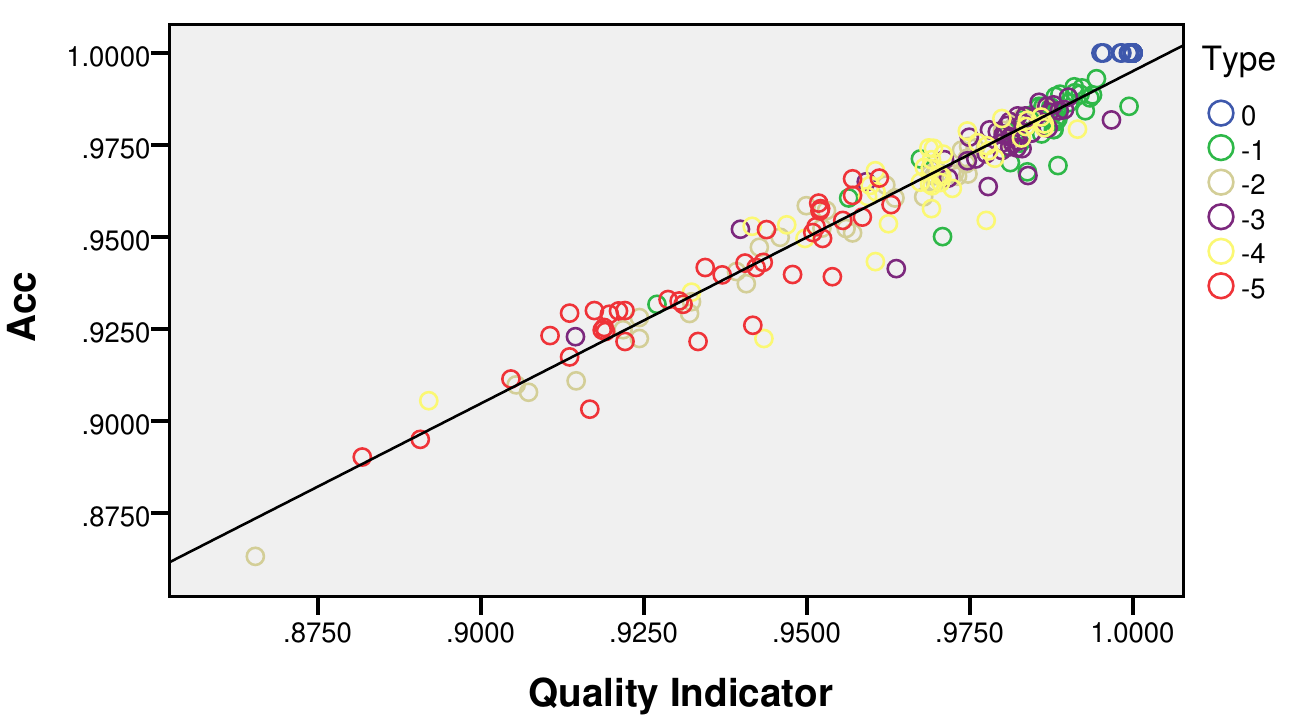}} %
	\subfloat[]{\includegraphics[width=0.5\textwidth]{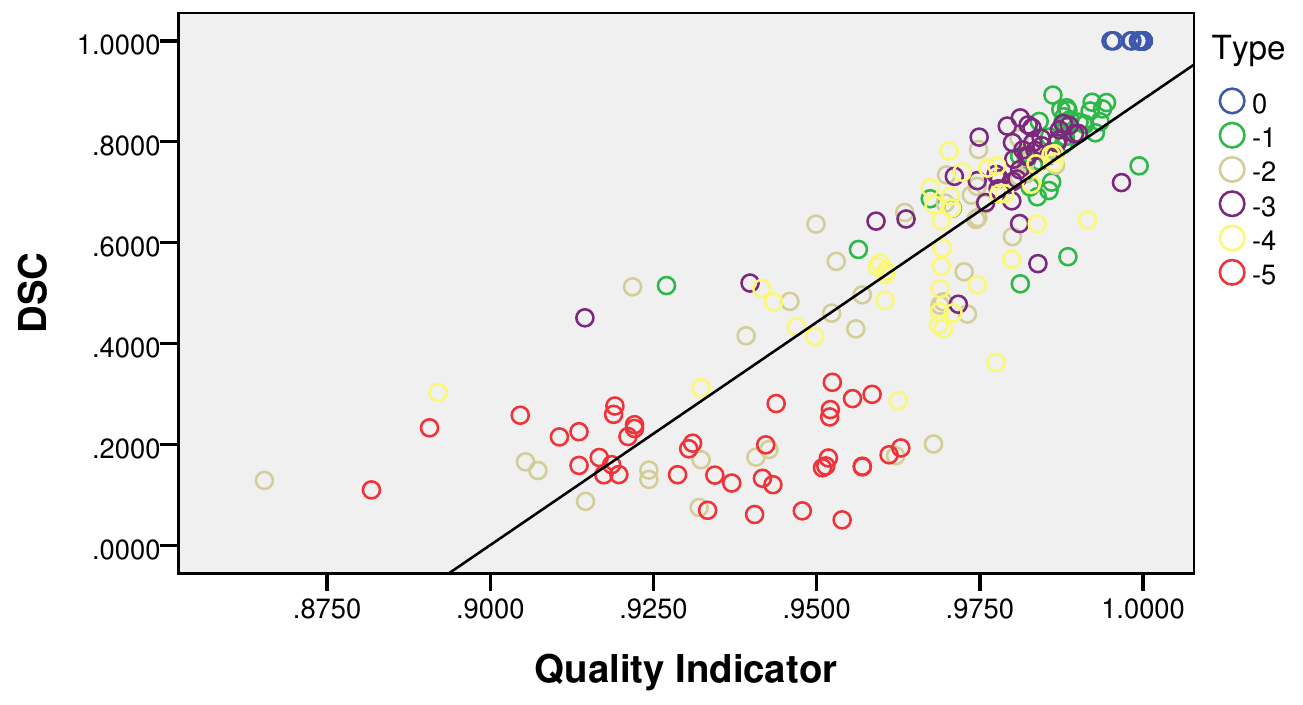}} %
	\caption{Scatter plots of (a) Real Acc and QI, and (b) Real DSC and QI. Different colors denotes different maks types (0: GT; -1: final output; -2 to -5: side outputs).} \label{fig_scatter}
\end{figure}

As mentioned in Section 2.3, we derive a QI by averaging the predicted error map. To measure how well QI can represent the segmentation quality, we compute the Pearson correlation coefficient (PCC), mean absolute error (MAE) between QI and real segmentation accuracy, as well as the PCC between QI and real DSC using all 220 masks, and the results are as follows:
\begin{equation}\label{eq_pcc}
PCC_{QI,Acc} = 0.972,\,PCC_{QI, DSC} = 0.856, \, MAE_{QI, Acc} = 0.0048,
\end{equation}
where both correlations are significant with $p<0.0001$. We also draw two scatter plots (QI-Acc and QI-DSC) in Fig. \ref{fig_scatter} to visualize the relationship between QI and the segmentation measurements. Considering the high PCC and low MAE between QI and Acc, as well as the strong linear relationship observed in Fig. \ref{fig_scatter}(a), QI can be regarded as a precise approximator to the real segmentation accuracy, thus can work as a good segmentation quality measurement.

\section{Conclusion}
Per-case and fine-grain segmentation quality assessment plays a crucial role in automating the image-based pipelines in medical research or clinical diagnosis, but this area has not yet been fully studied. This work formally defines a fine-grain error map prediction problem, and attempts to address it using a DL framework. The evaluation results on a public whole-heart segmentation dataset demonstrates the efficacy of our error map predictor, and also shows that a per-case quality measurement can be derived from the predicted error maps, which approximates the real segmentation accuracy well with a small MAE. Future work will investigate the potential of error map predictors in improving segmentor's robustness, where an error map predictor can serve as a critic to regularize the segmentation model. The proposed framework is inherently generic, in which the segmentors and predictors can be replaced with different types of models (e.g., random forest), and it can be adapted to other segmentation applications easily.

%
%
%

\end{document}